\pdfoutput=1
\documentclass[11pt]{article}

\usepackage[preprint]{acl}

\usepackage{times}
\usepackage{latexsym}
\usepackage{enumitem}
\usepackage{booktabs}

\usepackage[most]{tcolorbox}

\usepackage[T1]{fontenc}
\usepackage[utf8]{inputenc}

\usepackage{microtype}

\usepackage{inconsolata}

\usepackage{colortbl}
\usepackage{amsfonts,amssymb}       % blackboard math symbols
\usepackage{nicefrac}       % compact symbols for 1/2, etc.
\usepackage{microtype}      % microtypography
\usepackage{xcolor}         % colors
\usepackage{colortbl}
\usepackage{bigstrut}
\usepackage{pifont}% http://ctan.org/pkg/pifont
\usepackage{graphicx}
\usepackage{amsmath}
\usepackage{amssymb}
\usepackage{amsthm}
\usepackage{caption}
\usepackage{subfig}
\usepackage{multirow}
\usepackage{float}
\usepackage{wrapfig}
\usepackage{hyperref}

\title{MobiLlama: Towards Accurate and Lightweight Fully Transparent GPT} 
\author{%
  Omkar Thawakar$^{1}$\textsuperscript{\textnormal{*}}, Ashmal Vayani$^{1}$\textsuperscript{\textnormal{*}}, Salman Khan$^{1,2}$, \textbf{Hisham Cholakal}$^{1}$, \\
  \textbf{Rao M. Anwer}$^{1,3}$, \textbf{Michael Felsberg}$^{5}$, \textbf{Tim Baldwin}$^{1,4}$, \textbf{Eric P. Xing}$^{1}$, \textbf{Fahad Shahbaz Khan}$^{1,5}$\\
  [0.25cm]
  {\fontsize{10.5pt}{12pt}\selectfont $^{1}$Mohamed bin Zayed University of AI, $^{2}$Australian National University, $^{3}$Aalto University}\\
 {\fontsize{10.5pt}{12pt}\selectfont $^{4}$The University of Melbourne, $^{5}$Linköping University}\\
}

\begin{document}
\maketitle

\begin{abstract}
`\emph{Bigger the better}' has been the predominant trend in recent Large Language Models (LLMs) development.
However, LLMs do not suit well for scenarios that require on-device processing, energy efficiency, low memory footprint, and response efficiency. These requisites are crucial for privacy, security, and sustainable deployment. 
This paper explores the `\emph{less is more}' paradigm by addressing the challenge of designing accurate yet efficient Small Language Models (SLMs) for resource constrained devices. 
Our primary contribution is the introduction of an accurate and fully transparent open-source 0.5 billion (0.5B) parameter SLM, named \emph{MobiLlama}, catering to the specific needs of resource-constrained computing with an emphasis on enhanced performance with reduced resource demands.
\emph{MobiLlama} is a SLM design that initiates from a larger model and applies a careful parameter sharing scheme to reduce both the pre-training and the deployment cost.
%memory footprint during pre-training. 
%In contrast to existing approaches that scale down LLMs directly or via post-processing steps, MobiLlama is a novel SLM design that initiates from a larger 1.2B model and applies a careful parameter sharing scheme to reduce memory footprint during training. 
Our work strives to not only bridge the gap in open-source SLMs but also ensures full transparency, where complete training data pipeline, training code, model weights, and over 300 checkpoints along with evaluation codes is available at : \url{https://github.com/mbzuai-oryx/MobiLlama}.\def\thefootnote{*}\footnotetext{Equal contribution.}
\end{abstract}

\section{Introduction}
Recent years have witnessed a tremendous surge in the development of Large Language Models (LLMs) with the emergence of prominent closed-source commercial models such as ChatGPT, Bard, and Claude. These LLMs exhibit surprising capabilities, typically called emergent abilities, towards solving complex tasks. Most existing popular LLMs follow a similar trend that bigger is always better, where scaling model size or data size typically provides improved model capacity and performance on downstream tasks. For instance, the recent Llama-2 70 billion (70B) model~\cite{llama2} is considered more favorable in different chat applications due to its effectiveness towards handling dialogues, logical reasoning, coding, compared to its 7B counterpart which is typically better suited for basic tasks such as categorization or summaries. While these LLMs demonstrate impressive performance in handling complex language tasks, a key limitation is their size and computational requirements. For instance, the large-scale Falcon~\cite{falcon_tii} 180B model was trained using 4096 A100 GPUs and requires large memory and compute for deployment with dedicated high-performance servers and scalable storage systems.

Recently, Small Language Models (SLMs) have shown potential in terms of providing decent performance with emergent abilities achieved at a significantly smaller scale compared to their large-scale LLM counterparts. Modern SLMs like Microsoft's Phi-2 2.7 billion~\cite{phi2} highlight the growing focus in the community on achieving more with less. SLMs offer advantages in terms of efficiency, cost, flexibility, and customizability. With fewer parameters, SLMs offer significant computational efficiency in terms of fast pre-training and inference with reduced memory and storage requirements. This is critical in real-world applications where efficient resource utilization is highly desired. It particularly opens up possibilities in resource-constrained computing, where the models are required to be memory efficient to operate on low-powered devices (e.g., edge). SLMs support on-device processing that enhances privacy, security, response time, and personalization. Such an integration can lead to advanced personal assistants, cloud-independent applications, and improved energy efficiency with a reduced carbon footprint. 

The landscape of language models, especially SLMs, is currently marked by a notable lack of open-source availability. While LLMs have garnered significant attention, the proprietary nature of most models has led to limited transparency and accessibility, particularly in the realm of SLMs. This gap hinders the scientific and technological exploration of these more efficient, compact and performant models. Recognizing this, there's a growing need in the community for fully transparent open-source  SLMs, which would facilitate a deeper understanding of their capabilities and limitations and spur innovation by allowing broader community access to their architecture and reproducible training methodologies. We argue that bridging this gap is crucial for democratizing access to 
%advanced natural language processing (NLP) technologies and 
collaborative advancement for SLMs. Therefore, we investigate the problem of designing accurate yet efficient SLMs from scratch with the intention to provide full transparency in the form of access to entire training data pipeline and code, model weights, more than 300 checkpoints along with evaluation codes. 

%(e.g., 0.5 billion) specifically for edge and mobile devices

When designing a SLM from scratch it is desired that the resulting model is accurate, while maintaining efficiency 
in terms of pre-training and deployment.
%in terms of on-device memory and storage requirements.
%efficient in terms of on-device memory and storage requirements, while being accurate. 
A straightforward way is to scale-down a larger LLM design to the desired model size (e.g., 0.5B) by reducing either the size of the hidden dimension layers or the number of layers. We empirically observe both these design strategies to provide inferior performance. This motivates us to look into an alternative way of designing a SLM from scratch that is accurate yet maintains the efficiency,
%and maintains efficiency in terms of on-device memory and storage requirements, 
while offering full transparency.

\noindent\textbf{Contributions:} 

We introduce a SLM framework, named \textit{MobiLlama}, with an aim to develop accurate SLMs by alleviating the redundancy in the transformer blocks. Different to the conventional SLM design where dedicated feed forward layers (FFN) are typically allocated to each transformer block, we propose to employ a shared FFN design for all the transformer blocks within SLM. Our \emph{MobiLlama} leveraging a shared FFN-based SLM design is accurate and maintains efficiency, while offering full transparency in terms of data pipeline, training code, model weights and extensive intermediate checkpoints along with evaluation codes.

\begin{figure}[!t]
  \centering
    \includegraphics[width=0.96\linewidth]{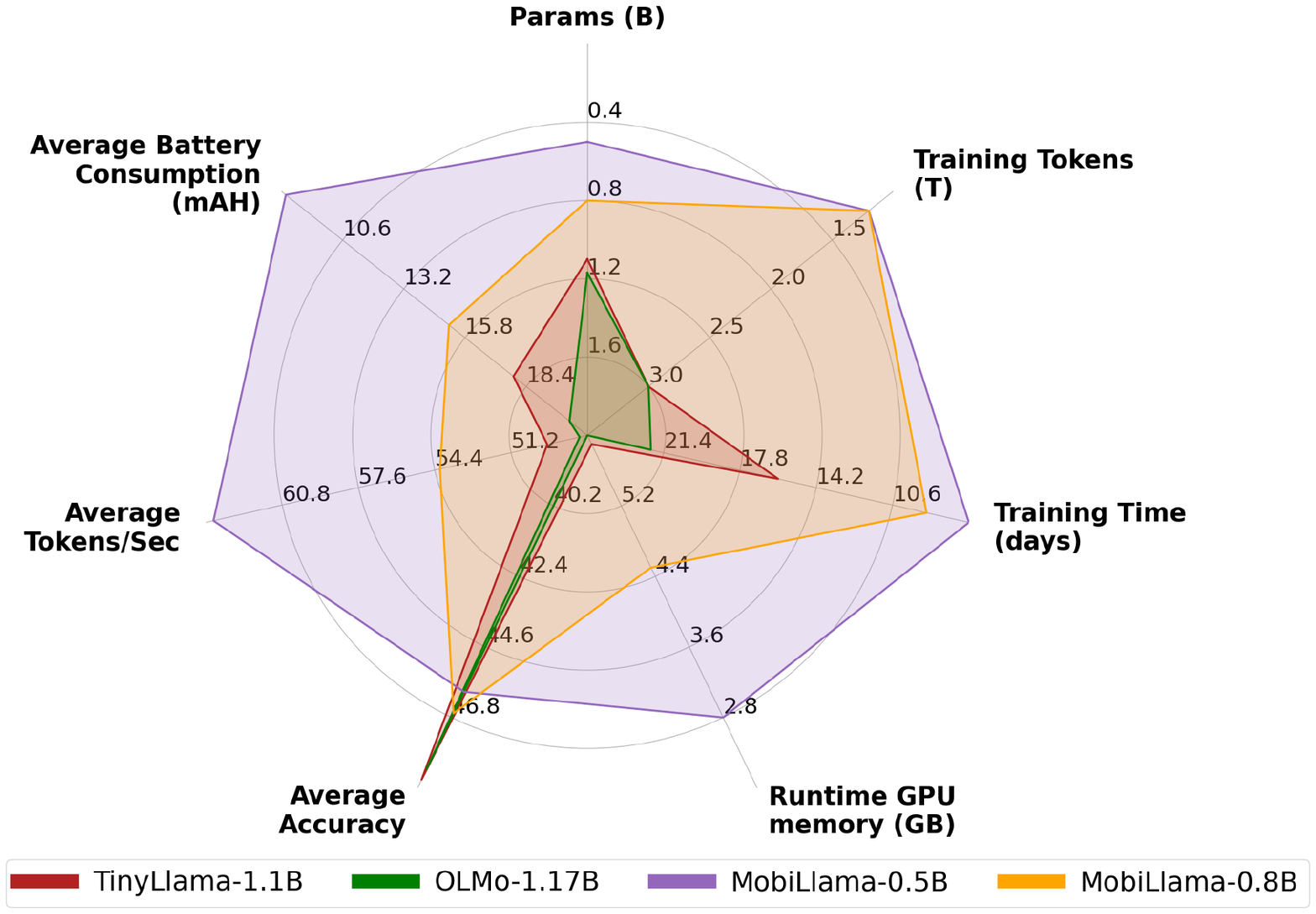}
    \vspace{-0.7em}
    \caption{
    Comparison of our \emph{MobiLlama}  0.5B and 0.8B models with recent OLMo-1.17B~\cite{OLMo} and TinyLlama-1.1B~\cite{tinyllama} in terms of pre-training tokens, pre-training time and memory, model parameters, overall accuracy across nine benchmarks and on-device efficiency (average battery consumption and average token/second on a PC with RTX2080Ti). Our  \emph{MobiLlama} achieves comparable accuracy while requiring significantly fewer pre-training data (1.2T tokens vs. 3T tokens), lesser pre-training time and GPU memory along with being efficient in terms of deployment on a resource constrained device. 
    }
    \label{fig:radar_plot}
    \vspace{-0.7em}
\end{figure}

We empirically show that our \emph{MobiLlama} performs favorably compared to conventional SLMs design schemes when performing pre-training from scratch. Our \emph{MobiLlama} 0.5B model outperforms existing SLMs of similar size on nine different benchmarks. \emph{MobiLlama} 0.5B achieves a gain of 2.4\% in terms of average performance on nine benchmarks, compared to the best existing 0.5B SLM in the literature. We further develop a 0.8B SLM that originates from our 0.5B model by utilizing a wider shared-FFN scheme in transformer blocks, achieving top performance %among existing $<$ 1B 
among existing SLMs falling under less than 1B parameters category.
%, with an average score of 45.12 on nine benchmarks. 
Lastly, we build multimodal models on top of our SLM to showcase visual perception and
reasoning capabilities. Fig.~\ref{fig:radar_plot} shows a comparison of our \emph{MobiLlama} with recent fully transparent relatively larger SLMs in terms of accuracy, pre-training complexity and on-board deployment cost.

\begin{table*}[!t]
\centering
{\small
\begin{tabular}{l|c|ccccc}
\toprule
Model & \#Params & Training Time & GPU Hours & GPU memory & No. of layers & Hidden dim size \\
\midrule
\emph{baseline1} & 0.54B & 7.5 days & 28.8K & 3.2 GB & 22 & 1024  \\
\emph{baseline2} & 0.52B & 7 days & 26.9K & 3 GB & 8 & 2048  \\
\emph{large-base} & 1.2B & 12 days & 46.1K & 6 GB & 22 & 2048  \\
\rowcolor{orange!15}\emph{MobiLlama} & 0.52B & 7 days & 26.6K & 3 GB & 22 & 2048  \\
\bottomrule
\end{tabular}}
\vspace{-0.5em}
\caption{Comparison of our \emph{MobiLlama} with the two baselines and the large-base model. We show the comparison in terms of total number of parameters, training time, total GPU hours, GPU memory, number of transformer layers and the hidden dimension size in each layer. The numbers are computed on A100 GPUs with 80 GB memory each. Compared to \emph{large-base}, our \emph{MobiLlama} reduces the GPU training hours by 42\% along with a significant reduction in GPU memory with the same design configuration (number of layers and hidden dimension size etc.). Further, our \emph{MobiLlama} possesses increased model capacity in terms of number of layers and hidden dimension size while maintaining comparable training cost and parameters, compared to \emph{baseline1} and \emph{baseline2}.}
%Compared to \emph{baseline1} and \emph{baseline2}}
\label{tab:baseline_comparison}
\vspace{-1.2em}
\end{table*}

\section{Related Work}
While LLMs have gained tremendous popularity~\cite{zhao2023survey}, one of their key limitations is the size and computational requirements both during pre-training and deployment. Another issue is limited availability of fully transparent opens-source LLMs that provide complete access to data pipeline, training code along with checkpoints and evaluation protocols. Prior works  explore making several components of LLM framework efficient such as, attention mechanism~\cite{dao2023flashattention} and optimization strategies~\cite{adamw}. Further, existing efforts also include exploring post-training sparsification schemes~\cite{slicegpt} or quantization~\cite{hoefler2021sparsity,zhu2023survey,xiao2023smoothquant} of computationally expensive LLM. In several cases, such a post-hoc sparsification can reduce the performance of LLMs with more on-device memory consumption, compared to a SLM trained from scratch. Further, these techniques typically employ LLMs with limited transparency and accessibility.

Recently, designing SLMs from scratch have gained attention~\cite{biderman2023pythia,lamini-lm,tinyllama,li2023starcoder,xglm,shoeybi2019megatron,zhang2022opt}. SLMs have shown potential as an alternative especially in case of limited pre-training compute as well as deployment in resource-constrained environments (e.g., edge devices). Further, SLMs can support on-device processing which in turn can enhance security, privacy, response efficiency, and personalization. Here, we strive to construct fully transparent accurate yet computationally efficient SLMs by maintaining the model's capacity to capture complex patterns and relationships in data while reducing the redundancy often present in the parameters of SLMs. Prior works~\cite{frantar2022gptq,gholami2022survey,pires2023one,pan2023interpretability,bhojanapalli2021leveraging} exploring alleviating redundancy in transformer design either focusing on the attention mechanism or on the single feed-forward layer in BERT style architectures. Different from these approaches, we explore alleviating the redundancy in the SLM architectures with an LLM objective function by focusing on the sharing mechanism of MLP blocks having multiple feed-forward network (FFN) layers. 

\section{Method}

\subsection{Baseline SLM Design}
\label{sec:baseline_design}

We first describe our baseline 0.5B SLM architecture that is adapted from recent TinyLlama~\cite{tinyllama} and  Llama-2~\cite{llama2}. The baseline architecture comprises $N$ layers, where each layer consists of hidden dimensions of $M$ and intermediate size (MLPs) of $5632$. The vocabulary size is 32$K$ and max. context length is $C$. We consider two different design choices when constructing a 0.5B model from scratch. In first design choice, named $baseline 1$, the number of layer is set to $N = 22$ and hidden size of each layer is set to $M = 1024$. In second design choice, named $baseline 2$, we set the number of layer to $N = 8$ and hidden size of each layer is set to $M = 2048$.

We note that both the aforementioned baseline designs struggle to strike an optimal balance between accuracy and efficiency. 
%Compared to the baseline 2, reducing the size of hidden dimensions to 1024 aids in computational efficiency. However, this can likely 
While a reduced size of hidden dimensions (1024) in case of $baseline 1$ aids in computational efficiency, it can likely hamper the model's capacity to capture complex patterns within the data. Such a reduction in dimension can potentially lead to a bottleneck effect, where the model's ability to represent intricate relationships and nuances in the data is constrained, thereby affecting the overall accuracy. On the other hand, reducing the number of hidden layers (22 to 8), as in the $baseline 2$, affects the model's depth  that in turn hampers its ability to learn hierarchical representations of the language. Achieving superior performance on tasks requiring deeper linguistic comprehension and contextual analysis likely requires combining the advantages of the two aforementioned baselines. However, increasing the model capacity of $baseline 1$ and $baseline 2$ into a single model (22 layers and hidden dimension size of 2048) results in a significantly larger parameterized model of 1.2B with increased training cost (see Tab.~\ref{tab:baseline_comparison}). We name this larger model as \emph{large-base}. Next, we present our proposed \emph{MobiLlama} 0.5B model design that does not reduce hidden dimension size in each layer ($baseline 1$) or the total number of layers ($baseline 2$), while maintaining a comparable training efficiency (see Tab.~\ref{tab:baseline_comparison}).

\begin{figure*}[t]
  \centering
    \includegraphics[width=0.96\linewidth]{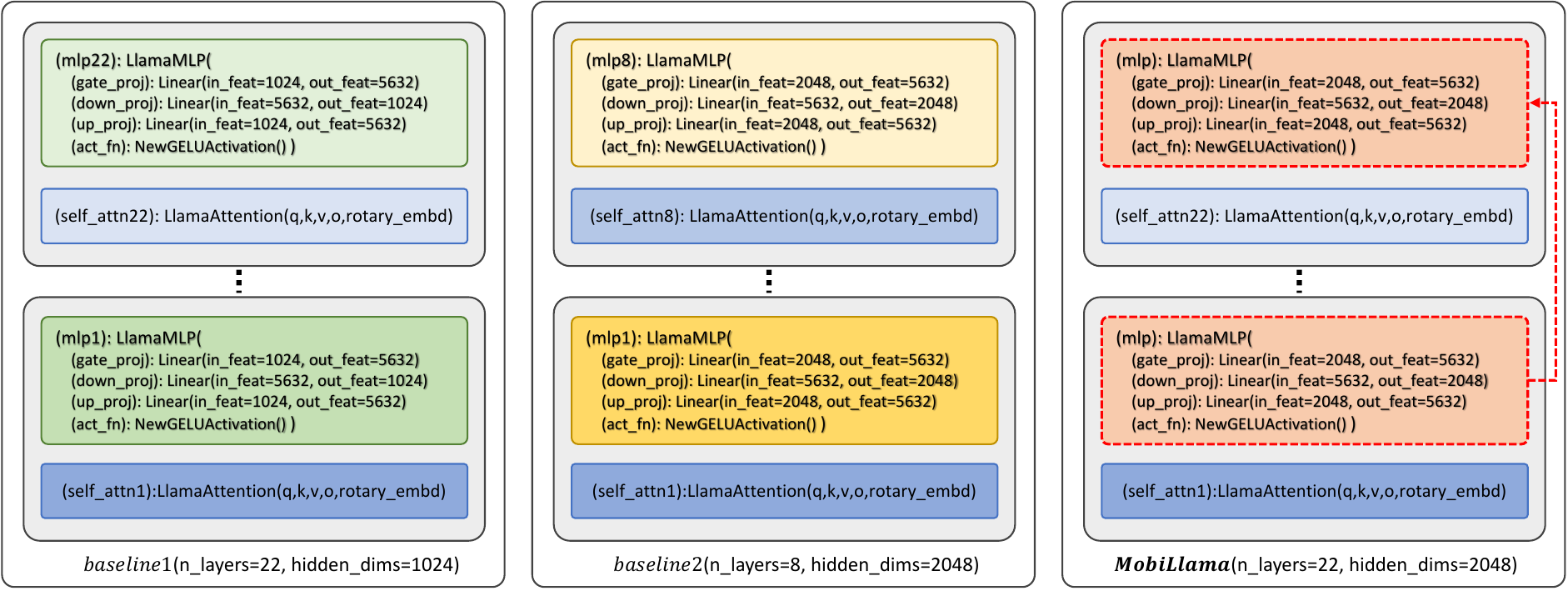}
    \caption{
    Illustrative comparison of our \emph{MobiLlama} with the two baselines. For each case, we show two transformer blocks denoted by different self-attention layers. In the case of both \emph{baseline1} and \emph{baseline2}, a dedicated MLP block comprising three FFN layers is utilized for each transformer layer. In contrast, our \emph{MobiLlama} utilizes a single MLP block (highlighted by the same color) that is shared across different transformer layers. This enables to increase the capacity of the network in terms of layers and hidden dimension size without any significant increase in the total number of trainable parameters.
    }
    \label{fig:Architecture}
\vspace{-1.5em}
\end{figure*}

\subsection{Proposed SLM Design: MobiLlama} 

The proposed approach, \emph{MobiLlama}, constructs a SLM of desired sizes (e.g., 0.5B model) by first initiating from a larger model size design, \emph{large-base}. Then, we employ a careful parameter sharing scheme to reduce the model size to a pre-defined model configuration, thereby significantly reducing the training cost. 
Generally, both SLMs and LLMs typically utilize a dedicated multilayer perceptron (MLP) block comprising multiple feed forward network (FFN) layers within each transformer block. In such a configuration (e.g., \emph{large-base}), the FFN layers account for a substantial $65\%$ of the total trainable parameters, with attention mechanisms and heads contributing $30\%$ and $5\%$, respectively. As a consequence, a significant number of parameters are concentrated within the FFN layers, thereby posing challenges during pre-training  with respect to computational cost and the model's ability to achieve faster convergence. To address these issues, we propose to use a sharing scheme where the FFN parameters are shared across all transformer layers within the SLM. This enables us to significantly reduce the overall trainable parameters by $60\%$ in our \emph{MobiLlama}, compared to the \emph{large-base}. Such a significant parameter reduction also enables us to increase the model capacity in terms of number of layers and hidden dimension size without any substantial increase in the training cost (see Tab.~\ref{tab:baseline_comparison}).  

Fig.~\ref{fig:Architecture} compares our architecture design with two baselines. In case of both baselines, a dedicated MLP block that consists of multiple FFN layers is used in each transformer layer. Instead, our efficient \emph{MobiLlama} design utilizes a single MLP block which is shared across different layers of transformer within the SLM. This helps in increasing the model capacity without any increase in the total number of trainable parameters in the model.

\begin{table}[H]
\centering
{\small
\begin{tabular}{lc}
\toprule
Subset & Tokens (Billion) \\ 
\midrule
Arxiv         & 30.00  \\
Book          & 28.86  \\
C4            & 197.67  \\
Refined-Web   & 665.01  \\
StarCoder     & 291.92  \\
StackExchange & 21.75   \\
Wikipedia     & 23.90   \\ 
\midrule
Total         & 1259.13 \\ 
\bottomrule
\end{tabular}}
% \vspace{-0.5em}
\caption{Data mix in Amber-Dataset.}
\label{tab:data_mix}
\vspace{-1.0em}
\end{table}

\begin{table}[H]
\centering
{\small
\begin{tabular}{ll}
\toprule
Hyperparameter & Value \\
\midrule
Number Parameters & 0.5$B$\\
Hidden Size & 2048 \\
Intermediate Size (in MLPs) & 5632 \\
Number of Attention Heads & 32 \\
Number of Hidden Layers & 22 \\
RMSNorm $\epsilon$ & $1e^{-6}$ \\
Max Seq Length & 2048 \\
Vocab Size & 32000 \\
\bottomrule
\end{tabular}}
% \vspace{-0.5em}
\caption{\emph{MobiLlama} architecture \& hyperparameters.}
\label{tab:mobillama-config}
% \vspace{-1.2em}
\end{table}

\subsection{Towards Fully Transparent MobiLlama}

\begin{table*}[!t]
\setlength{\tabcolsep}{2pt}
\centering
{
\small
\begin{tabular}{l|c|ccccccccc|c}
\toprule
Model Name & \#Params & HellaSwag & Truthfulqa & MMLU & Arc\_C & CrowsPairs & piqa & race & siqa & winogrande & Average \\
\midrule
gpt-neo-125m & 0.15B & 30.26 & 45.58 & 25.97 & 22.95 & 61.55 & 62.46 & 27.56 & 40.33 & 51.78 & 40.93  \\
tiny-starcoder & 0.17B & 28.17 & 47.68 & 26.79 & 20.99 & 49.68 & 52.55 & 25.45 & 38.28 & 51.22 & 37.86  \\
cerebras-gpt-256m & 0.26B & 28.99 & 45.98 & 26.83 & 22.01 & 60.52 & 61.42 & 27.46 & 40.53 & 52.49 & 40.69  \\
opt-350m & 0.35b & 36.73 & 40.83 & 26.02 & 23.55 & 64.12 & 64.74 & 29.85 & 41.55 & 52.64 & 42.22 \\
megatron-gpt2-345m & 0.38B & 39.18 & 41.51 & 24.32 & 24.23 & 64.82 & 66.87 & 31.19 & 40.28 & 52.96 & 42.81  \\
LiteLlama & 0.46B & 38.47 & 41.59 & 26.17 & 24.91 & 62.90 & 67.73 & 28.42 & 40.27 & 49.88 & 42.26  \\
gpt-sw3-356m & 0.47B & 37.05 & 42.55 & 25.93 & 23.63 & 61.59 & 64.85 & 32.15 & 41.56 & 53.04 & 42.48  \\
pythia-410m & 0.51B & 40.85 & 41.22 & 27.25 & 26.19 & 64.20 & 67.19 & 30.71 & 41.40 & 53.12 & 43.57  \\
xglm-564m & 0.56B & 34.64 & 40.43 & 25.18 & 24.57 & 62.25 & 64.85 & 29.28 & 42.68 & 53.03 & 41.87  \\
Lamini-GPT-LM & 0.59B & 31.55 & 40.72 & 25.53 & 24.23 & 63.09 & 63.87 & 29.95 & 40.78 & 47.75 & 40.83  \\
\rowcolor{orange!15} \textbf{MobiLlama (Ours)} & 0.5B & 52.52 & 38.05 & 26.45 & 29.52 & 64.03 & 72.03 & 33.68 & 40.22 & 57.53 & \textbf{46.00} \bigstrut \\
\midrule
Lamini-GPT-LM & 0.77B & 43.83 & 40.25 & 26.24 & 27.55 & 66.12 & 69.31 & 37.12 & 42.47 & 56.59 & 45.49  \\
\rowcolor{orange!15} \textbf{MobiLlama (Ours)} & 0.8B & 54.09 & 38.48 & 26.92 & 30.20 & 64.82 & 73.17 & 33.37 & 41.60 & 57.45 & \textbf{46.67} \\
\bottomrule
\end{tabular}
}
\caption{State-of-the-art comparisons with existing \textit{< 1B params models} on \textit{nine} benchmarks. In case of around 0.5B model series, our \emph{MobiLlama} achieves a substantial gain of {2.4\%} in terms of average performance on nine benchmarks. Further, our \emph{MobiLlama} 0.8B model achieves an average score of 46.67. }
\label{tab:llm-comparison}
\vspace{-1.2em}
\end{table*}

As discussed earlier, fully transparent open-source SLM development is desired to foster a more inclusive, data/model provenance, and reproducible collaborative SLM research development environment. To this end, we present here pre-training dataset and processing details, architecture design configuration with training details, evaluation benchmarks and metrics. In addition, we will publicly release complete training and evaluation codes along with intermediate model checkpoints.

\noindent\textbf{Pre-training Dataset and Processing:}
For pre-training, we use 1.2T tokens from  LLM360 Amber dataset~\cite{llm360}. The Amber dataset  provides a rich and varied linguistic landscape having different text types, topics, and styles. 
Tab.~\ref{tab:data_mix} shows the data mix from Amber dataset gathered from various sources.

\textit{Arxiv (30 Billion Tokens)} subset is drawn from the repository of scientific papers, provides complex, domain-specific language and technical terminology, enriching the understanding of academic prose. \textit{Book (28.9 Billion Tokens)} subset comprises tokens from a broad range of literature with diverse narrative styles, cultural contexts, and rich vocabulary, deepening the grasp of storytelling and language nuances. \textit{C4 (197.7 Billion Tokens)} is the Colossal Clean Crawled Corpus (C4) that offers a vast and cleaned selection of web text, providing a broad linguistic foundation that includes various registers, styles, and topics. \textit{Refined-Web (665 Billion Tokens)} subset is a curated web crawl and offers the model exposure to contemporary, informal, and varied internet language, enhancing the relevance and applicability to modern communication. \textit{StarCoder (291.9 Billion Tokens)} subset is a vast collection used for code understanding featuring 783GB of code across 86 programming languages. It includes GitHub issues, Jupyter notebooks, and commits, totaling approximately 250 billion tokens. These are meticulously cleaned and de-duplicated for training efficiency. \textit{StackExchange (21.8 Billion Tokens)} is from the network of Q\&A websites, this subset aids the model in learning question-answering formats and technical discussions across diverse topics. \textit{Wikipedia (23.9 Billion Tokens)} is an encyclopedia collection, it offers well-structured and factual content that helps the model to learn encyclopedic knowledge and formal writing styles.

From the above-mentioned subsets, Arxiv, Book, C4, StackExchange and Wikipedia are sourced from RedPajama-v1~\cite{together2023redpajama}. The Amber dataset uses RefinedWeb~\cite{refinedweb} data to replace common\_crawl subset of RedPajama-v1. These subsets amount to 1259.13 billion tokens.

Initially, raw data sourced from the above sources is tokenized using Huggingface LLaMA tokenizer~\cite{llama2}. Subsequently, these tokens are organized into sequences with each containing 2048 tokens. To manage data, these sequences are merged to the token sequences and divided the amalgamated dataset into 360 distinct segments. Each data segment, structured as a jsonl file, carries an array of token IDs along with a source identifier that denotes the originating dataset. Each data sample is designed to have 2049 tokens.

\noindent \textbf{Architecture Design:}
Our \emph{MobiLlama} 0.5B comprises a hidden size of 2048, an intermediate size of 5632 in its MLPs, and operates with 32 attention heads across 22 hidden layers. It is designed to handle sequences up to 2048 tokens long, supported by a vocabulary size of 32,000. The precision in normalization is ensured by an RMSNorm epsilon of $1e^{-6}$ to obtain a more stable training. We utilize RoPE (Rotary Positional Embedding)~\cite{su2024roformer} to encode positional information in our \emph{MobiLlama}. Similar to \cite{tinyllama}, we employ a combination of Swish and Gated Linear Units together as activation functions.  
Tab.~\ref{tab:mobillama-config} presents details of our model configuration. 
We also derive a 0.8B version from our \emph{MobiLlama} by widening the shared FFN design. Compared to the 0.5B model, our 0.8B design increases the hidden dimension size to 2532 and the intermediate size to 11,080 while the rest of the configuration is same.

\begin{table}[!t]
\setlength{\tabcolsep}{1pt}
\centering
{
\small
\begin{tabular}{l|cccc|c}
\toprule
Model & HellaSwag & Truthfulqa & MMLU & Arc\_C & Average  \\
\midrule
\emph{baseline1} & 42.44 & 38.46 & 25.08 & 26.18 & 33.04  \\
\emph{baseline2} & 42.15 & 38.70 & 25.73 & 26.10 & 33.17  \\
\rowcolor{orange!15} \emph{MobiLlama} & 44.47 & 40.12 & 26.48 & 26.53 & \textbf{34.40}  \\
\bottomrule
\end{tabular}
\vspace{-1em}
}
\caption{Baseline comparison on four benchmarks. Here, both the baselines and our \emph{MobiLlama}  comprise the same parameters (0.5B) and are pre-trained on 100B tokens from  Amber. Our \emph{MobiLlama}  achieves favorable performance compared to the two baselines, while operating on a similar training budget. 
}
\label{tab:basemodel_selection}
\vspace{-1.2em}
\end{table}

\begin{table*}[!t]
\setlength{\tabcolsep}{4pt}
\centering
{
\resizebox{\textwidth}{!}{
\begin{tabular}{c l c c cccc}
\toprule
\multirow{2}{*}{Platform}& \multirow{2}{*}{Model} & \multirow{1}{*}{\#Params} & \multirow{2}{*}{Precision} & {Avg Tokens/Sec} & Avg Memory   & Avg Battery Consumption & CPU   \\
& &   ($\downarrow$) &  & ($\uparrow$) & Consumption ($\downarrow$) & /1k Tokens ($\downarrow$) & Utilization ($\downarrow$)   \\
\midrule
\multirow{4}{*}{RTX2080Ti} & Llama2 & 7B & bf16 & 14.85 & 27793 MB & 135.51 mAH & 31.62\% \\
& Phi2  & 2.7B & bf16 & 32.19 & 12071 MB & 59.13 mAH  & 24.73\% \\
& \emph{large-base} & 1.2B & bf16 & 50.61 & 6254 MB & 18.91 mAH  & 18.25\%   \\
 \rowcolor{blue!10} \cellcolor{white} &   \emph{MobiLlama}  & 0.5B & bf16 & \textbf{63.38} & \textbf{3046} MB & \textbf{8.19} mAH  & \textbf{14.79}\% \\
\midrule
\multirow{4}{*}{CPU-i7} & Llama2 & 7B & 4bit & 5.96 & 4188 MB & 73.5 mAH & 49.16\%   \\
& Phi2 & 2.7B & 4bit & 22.14 & 1972 MB & 27.36 mAH  & 34.92\%   \\
& \emph{large-base}   & 1.2B & 4bit & 29.23 & 1163 MB & 10.81 mAH  & 30.84\%   \\
\rowcolor{blue!10} \cellcolor{white} &  \emph{MobiLlama}  & 0.5B & 4bit & \textbf{36.32} & \textbf{799} MB & \textbf{4.86} mAH  & \textbf{24.64}\%   \\
\midrule
\multirow{4}{*}{Snapdragon-685} & Llama2 & 7B & 4bit & 1.193 & 4287 MB & 10.07 mAH & 77.41\%   \\
& Phi2  & 2.7B & 4bit & 2.882 & 1893 MB & 14.61 mAH  & 56.82\%   \\
& \emph{large-base} & 1.2B & 4bit & 6.687 & 780 MB & 6.00 mAH  & 17.15\%  \\
 \rowcolor{blue!10} \cellcolor{white} & \emph{MobiLlama}  & 0.5B & 4bit & \textbf{7.021} & \textbf{770} MB & \textbf{5.32} mAH  & \textbf{13.02}\% \\
\bottomrule
\end{tabular}}
}
\vspace{-0.5em}
\caption{Comparison in terms of efficiency and resource consumption on different low-end hardware devices. We show the comparison on: a PC with RTX-2080Ti GPU, a laptop with i7 CPU and a smartphone with Snapdragon-685 processor. In addition to our \emph{large-base} model, we also present the comparison with Llama2 7B and Phi2 2.7B. In case of CPU and smartphone, we use 4-bit GGUF format of the corresponding models, whereas the original models are deployed and tested on PC with RTX-2080Ti GPU. The different metrics measure the model's operational efficiency, model's footprint in the device's RAM and the energy efficiency of processing 1,000 tokens. Our \emph{MobiLlama} performs favorably in terms of efficiency on these low-end hardware devices. We note that both Phi2 and Llama2 are not fully transparent in that the complete data pipeline for pre-training is not publicly available. 
}
\label{tab:on-device-comparison}
\vspace{-0.5em}
\end{table*}

% \begin{table*}[!t]
% \centering
% \begin{tabular}{clcccccc}
% \toprule
% Platform & Model & \#Params & Precision & Avg Tokens/Sec & Avg Memory Consumption & Avg Battery Consumption & CPU Utilization \\
% \midrule
% \multicolumn{8}{c}{RTX2080Ti} \\
% Llama2 & 7B & bf16 & 14.85 & 27793 MB & 135.51 mAH & 31.62\% \\
% Phi2 & 2.7B & bf16 & 32.19 & 12071 MB & 59.13 mAH & 24.73\% \\
% \emph{large-base} & 1.2B & bf16 & 50.61 & 6254 MB & 18.91 mAH & 18.25\% \\
% \emph{MobiLlama} & 0.5B & bf16 & 63.38 & 3046 MB & 8.19 mAH & 14.79\% \\
% \midrule
% \multicolumn{8}{c}{CPU-i7} \\
% ... & ... & ... & ... & ... & ... & ... & ... \\
% \bottomrule
% \end{tabular}
% \caption{Comparison in terms of efficiency and resource consumption on different low-end hardware devices.}
% \label{tab:on-device-comparison}
% \end{table*}

For pre-training of our \emph{MobiLlama}, we use a public cluster having 20 GPU nodes each equipped with 8 NVIDIA A100 GPUs with 80 GB memory each and 800 Gbps interconnect for model training. Each GPU is interconnected through 8 NVLink links, complemented by a cross-node connection configuration of 2 port 200 Gb/sec (4$\times$ HDR) InfiniBand, optimizing the model's training process. To further enhance the training efficiency, we employ flash-attention mechanism and follow the pre-training hyper-parameters established by the LLaMA~\cite{llama2} model. Our \emph{MobiLlama} model's training is performed using the AdamW optimizer, leveraging hyperparameters $\beta_1=0.9, \beta_2=0.95$, with an initial learning rate of $\eta=3e^{-4}$. This rate follows a cosine learning rate schedule, tapering to a final rate of $\eta=3e^{-5}$. We further incorporate a weight decay of $0.1$ and apply gradient clipping at $1.0$ with a warm-up period over $2,000$ steps. Adapting to our hardware configuration of 20 GPU nodes, we optimize the pre-training batch size to $800$ ($160 \times 5$), achieving a throughput of approximately 14$k$-15$k$ tokens per second on a single GPU. During our model pre-training, we save intermediate checkpoints after every 3.3B tokens which will be publicly released.

\noindent \textbf{Evaluation Benchmarks and Metrics:}

For a comprehensive performance evaluation, we use nine different benchmarks from the Open LLM Leaderboard\footnote{{\tiny\url{https://huggingface.co/spaces/HuggingFaceH4/open_llm_leaderboard}}}. 

HellaSwag~\cite{zellers2019hellaswag} assesses the model's ability to predict the correct ending to a scenario from a set of possible continuations, thereby testing common sense reasoning. TruthfulQA~\cite{lin2021truthfulqa} evaluates the model to provide truthful answers, focusing on its understanding of facts and its ability to avoid deception. MMLU~\cite{hendrycks2020measuring} measures the model's broad knowledge across numerous subjects such as, humanities, science, technology, engineering and management. ARC\_Challenge~\cite{clark2018think} tests complex reasoning with science questions. CrowsPairs~\cite{nangia2020crows} evaluates the model's biases by comparing sentences that differ only by the demographic group mentioned, aiming for fairness. PIQA~\cite{bisk2020piqa} evaluates the model's physical commonsense knowledge, requiring understanding of everyday physical processes. Race~\cite{lai2017race} assesses reading comprehension through multiple-choice questions based on passages. SIQA~\cite{sap2019socialiqa} focuses on the model's social commonsense reasoning and its understanding of social dynamics. Winogrande~\cite{sakaguchi2021winogrande} evaluates the model's ability to resolve ambiguities in text, testing its commonsense reasoning.

Following the Analysis-360 framework~\cite{llm360} that is built on llm-harness~\cite{llm_harness}, we conduct extensive evaluations under the standard settings with varying shots for detailed assessments, validating the model's robustness and adaptability across diverse linguistic tasks. Following the standard evaluation protocol, our evaluation setting consists of 10, 25, 5 and 5  shot evaluation for Hellaswag, ARC\_Challenge, Winogrande and MMLU, while zero-shot for rest of the benchmarks.

\section{Results}

\noindent\textbf{Baseline Comparison:}
We first present a comparison with the two baselines in Tab.~\ref{tab:basemodel_selection}) for 0.5B model series. For the baseline evaluation, we pre-train all the models on the same 100B tokens from the Amber dataset and report the results on four benchmarks: HellaSwag, TruthfulQA, MMLU, and Arc\_C. Our \emph{MobiLlama} achieves favourable performance compared to the two baselines by achieving an average score of 34.4 over the four benchmarks. We note that this performance improvement is achieved without any significant increase in the training cost (see Tab.~\ref{tab:baseline_comparison}), highlighting the merits of the proposed SLM design. \\
\noindent\textbf{State-of-the-art Comparison:}
We compare our \emph{MobiLlama}  0.5B and 0.8B with existing SLMs having comparable (less than 1B) parameters: gpt-neo~\cite{gpt-neo}, tiny-starcoder~\cite{li2023starcoder}, cerebras-gpt~\cite{dey2023cerebras}, opt~\cite{zhang2022opt}, megatron-gpt-2~\cite{shoeybi2019megatron}, LiteLlama, gpt-sw3, pythia~\cite{biderman2023pythia}, xglm~\cite{xglm}, Lamini-LM~\cite{lamini-lm}. 
Among existing methods falling around 0.5B model series category, pythia-410m achieves an average score of 43.57. Our \emph{MobiLlama} 0.5B model achieves superior performance with an average score of 46.0, outperforming pythia-410m by 2.4\% in terms of average performance on nine benchmarks. Notably, \emph{MobiLlama} achieves superior performance on the HellaSwag benchmark which is designed to evaluate the model's capabilities in the NLP text completion task. Further, \emph{MobiLlama} also performs favorably on commonsense reasoning tasks with superior results on piqa and winogrande benchmarks. Further, our \emph{MobiLlama} 0.8B model achieves an average score of 49.06.

\begin{table}[!t]
\setlength{\tabcolsep}{2pt}
\centering
{
\resizebox{\columnwidth}{!}{
\begin{tabular}{l | c c|cccc | c}
\toprule
Model  & \#Slice & \#Params & HellaS  & Arc\_C & piqa  & wino & Average  \\
\midrule
OPT-1.3B & 30\% & 0.91B & 39.81 & 25.77 & 60.77 & 54.7 & 45.26  \\
OPT-6.7B & 30\% & 4.69B & 54.56 & 29.01 & 68.61 & 60.69 & 53.21  \\
Llama-2-7B & 30\% & 4.9B & 49.62 & 31.23 & 63.55 & 61.33 & 51.43  \\
Phi2-2.7B & 30\% & 1.89B & 47.56 & 30.29 & 65.94 & 63.14 & 51.73  \\
\midrule
\rowcolor{orange!15}  & Dense & 0.5B & 52.52 & 29.52 & 72.03 & 57.53 & 52.90  \\
\rowcolor{orange!15} \multirow{-2}{*}{ \textbf{\emph{MobiLlama}} } & Dense & 0.8B & 54.09 & 30.20 & 73.17 & 57.45 & 53.72  \\
\bottomrule
\end{tabular}}
}
\vspace{-0.5em}
\caption{Comparison on \textit{4 open LLM benchmarks} when parameters are sliced down to 30\% using Wiki2Text dataset, following~\cite{slicegpt}.}
\label{tab:sliced-llm-comparison}
\vspace{-1em}
\end{table}
\begin{table}[!t]
\setlength{\tabcolsep}{6pt}
\centering
{
\small
\begin{tabular}{l cccccc}
\toprule
Model & GQA & SQA & TextQA & MME  
\\
\midrule

\emph{MobiLlama}-$V$ & 58.5 & 53.1 & 41.4 & 1191.9   
\\
\bottomrule
\end{tabular}
\vspace{-1em}
}
\caption{Quantitative performance of our  multimodal design, \emph{MobiLlama}-$V$ 0.8B, on different benchmarks. }
\label{tab:vlm-comparison}
\vspace{-1.4em}
\end{table}

\begin{table*}[t]
\setlength{\tabcolsep}{2pt}
\centering
{
\small
\begin{tabular}{l|c|ccccccccc|c}
\toprule
Model & \#Params & HellaSwag & Truthfulqa & MMLU & Arc\_C & CrowsPairs & piqa & race & siqa  & winogrande & Average  \\
\midrule
Boomer & 1B & 31.62 & 39.42 & 25.42 & 22.26 & 61.26 & 57.99 & 28.99 & 40.32 & 50.98 & 39.80  \\
Pythia-Dedup & 1B & 49.63 & 38.92 & 24.29 & 29.09 & 67.11 & 70.23 & 32.44 & 42.63 & 53.98 & 45.36  \\
Falcon-RW & 1B & 63.12 & 35.96 & 25.36 & 35.06 & 69.04 & 74.10 & 36.07 & 40.23 & 61.88 & 48.98  \\
TinyLlama & 1.1B & 60.22 & 37.59 & 26.11 & 33.61 & 70.60 & 73.28 & 36.45 & 41.65 & 59.18 & 48.74  \\
OLMo & 1.2B & 62.50 & 32.94 & 25.86 & 34.45 & 69.59 & 73.70 & 36.74 & 41.14 & 58.90 & 48.42  \\
Cerebras-GPT & 1.3B & 38.51 & 42.70 & 26.66 & 26.10 & 63.67 & 66.75 & 30.33 & 42.42 & 53.59 & 43.41  \\
Lamini & 1.3B & 38.05 & 36.43 & 28.47 & 26.62 & 64.62 & 67.89 & 33.39 & 43.19 & 50.59 & 43.25  \\
OPT & 1.3B & 54.50 & 38.67 & 24.63 & 29.6 & 70.70 & 72.47 & 34.16 & 42.47 & 59.74 & 47.43  \\
GPT-NEO & 1.3B & 48.49 & 39.61 & 24.82 & 31.31 & 65.67 & 71.05 & 34.06 & 41.81 & 57.06 & 45.98  \\
Pythia-Deduped & 1.4B & 55.00 & 38.63 & 25.45 & 32.59 & 67.33 & 72.68 & 34.64 & 42.68 & 56.90 & 47.32  \\
\rowcolor{orange!15} \textbf{\emph{large-base}} & 1.2B & 62.99 & 35.90 & 24.79 & 34.55 & 68.49 & 75.57 & 35.31 & 41.96 & 62.03 & \textbf{49.06}  \\
\bottomrule
\end{tabular}
}
\vspace{-0.5em}
\caption{Comprehensive comparisons with existing \textit{< 2B params fully open-source LLM models} on \textit{9} benchmarks. Our 1.2B \emph{large-base} model pre-trained on 1.2T tokens achieves superior performance compared to both the recent OLMo 1.17B model~\cite{OLMo} and TinyLlama 1.1B model~\cite{tinyllama}, which are pre-trained on a substantially larger data of 3T tokens.
}
\label{tab:llm-comparison-1B}
\vspace{-0.5em}
\end{table*}

\begin{figure*}[!t]
  \centering
    \includegraphics[width=0.96\linewidth]{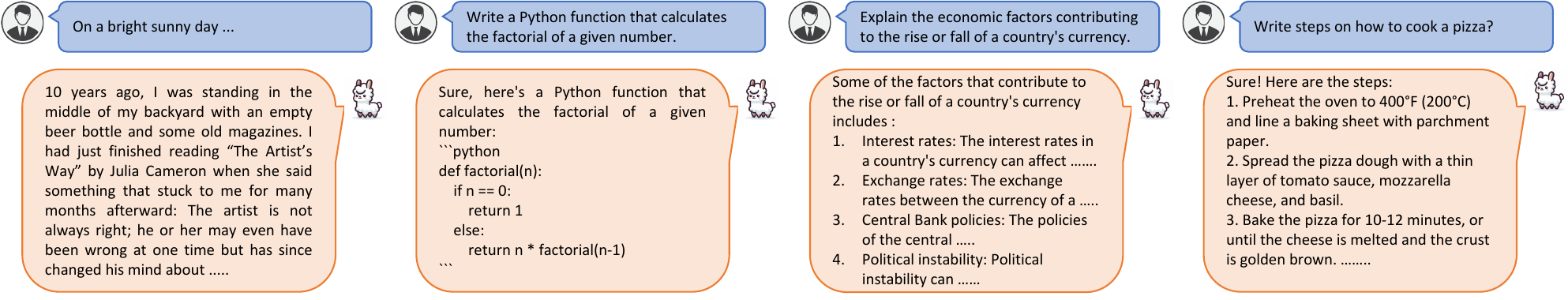}
    \vspace{-0.7em}
    \caption{
    Example responses from our \emph{MobiLlama} across a variety of tasks, including creative storytelling, coding exercises, economic analysis, and cooking instructions. The responses highlight the models' ability to engage with both abstract concepts and practical, step-by-step processes, demonstrating its broad applicability.
    }
    \label{fig:llm_examples}
    \vspace{-0.7em}
\end{figure*}
\begin{figure*}[!t]
  \centering
    \includegraphics[width=0.96\linewidth]{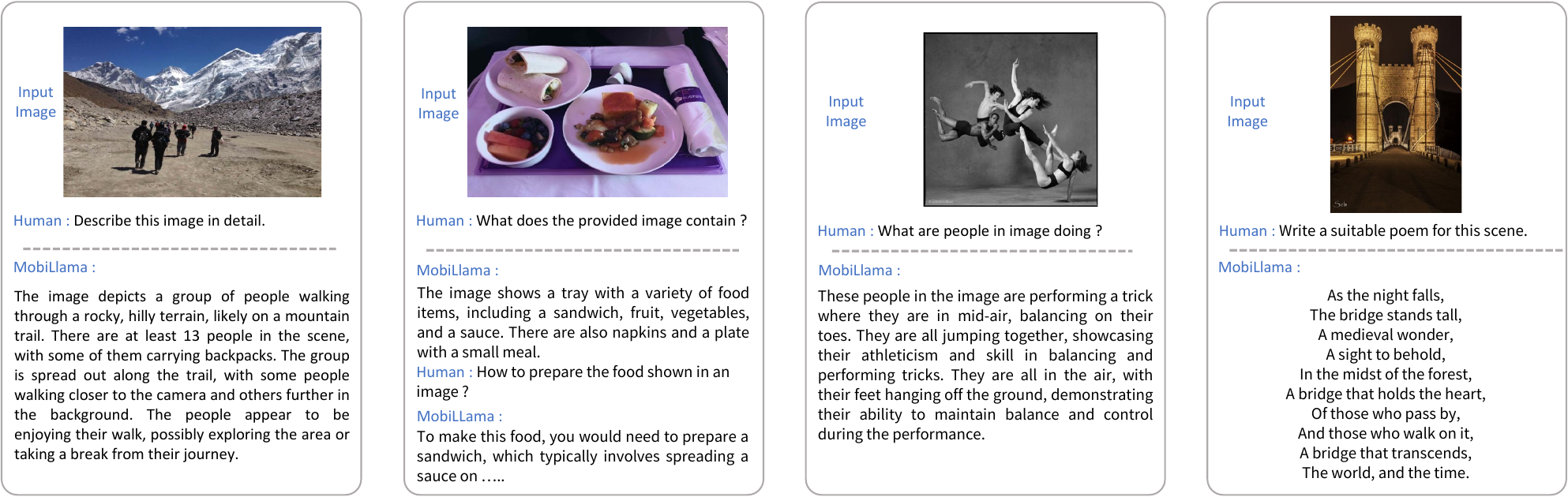}
    \vspace{-0.7em}
    \caption{
    Example responses of \emph{MobiLlama}-$V$ in responding to visual stimuli across a range of scenarios.
    }
    \label{fig:vlm_examples}
    \vspace{-1.0em}
\end{figure*}

\noindent\textbf{Efficiency Comparison:}
We present the comparison of our model in terms of efficiency and resource consumption on various low-end hardware platforms:  a PC with RTX-2080Ti GPU, a laptop with i7 CPU, and a smartphone with Snapdragon-685
processor. Tab.~\ref{tab:on-device-comparison}
shows the comparison of our \emph{MobiLlama} 0.5B with \emph{large-base} 1.2B, Llama2-7B~\cite{llama2} and Phi2-2.7B~\cite{phi2} model, in terms of the average processing speed in tokens per second (Average Tokens/Sec), average memory consumption (Avg Memory Consumption) in megabytes (MB), and the average battery consumption (Average Battery Consumption/1000 Tokens) in milliampere-hours (mAH). Our \emph{MobiLlama} performs favorably in terms of efficiency across different hardware platforms.

%in terms of its ability to process tokens more efficiently while utilizing less memory and CPU resources. 

We further perform an efficiency comparison to a recent post-training sparsification scheme~\cite{slicegpt}, where each weight matrix is substituted with a smaller (dense) matrix, thereby reducing dimensions of the embeddings in the model. In such a scheme, the parameters of the original LLM are reduced significantly up to 70\% followed by post-slicing fine-tuning using a dataset such as WikiText-2~\cite{wikitext}. Tab.~\ref{tab:sliced-llm-comparison} shows the comparison of our \emph{MobiLlama} with existing LLMs (e.g., Llama-2-7B, OPT-6.7B) on four benchmarks following~\cite{slicegpt}. Our \emph{MobiLlama} 0.5B and 0.8B models perform favorably against representative LLMs, with an average score of 53.72 computed over four benchmarks. These results highlight the potential of designing new fully transparent SLMs that can achieve comparable capabilities of their larger sliced model counterparts.

\noindent \textbf{Multimodal MobiLlama:}
We further build a multimodal model on top of our \emph{MobiLlama} by combining it with a vision encoder to develop a general-purpose visual assistant having visual reasoning capabilities. Our multimodal model, \emph{MobiLlama}-$V$, is trained by bridging the visual encoder of CLIP~\cite{radford2021learning} with the language decoder of our \emph{MobiLlama}, and fine-tuning
it in an end-to-end fashion on a 665k vision-language instruction set~\cite{liu2023llava}. We conduct evaluation on GQA~\cite{hudson2019gqa}, SQA~\cite{lu2022learn}, TextQA~\cite{singh2019towards}, and MME~\cite{fu2023mme}.
Tab.~\ref{tab:vlm-comparison} shows the performance of \emph{MobiLlama}-$V$ 0.8B model.

\noindent \textbf{Qualitative Analysis:} Fig.~\ref{fig:llm_examples} shows example responses obtained when interacting with \emph{MobiLlama} 0.5B with conversation capabilities. We show examples covering different tasks such as, text completion, code generation and conversation capabilities. Our model generates faithful responses to these diverse interactions. 
%such as, asking to generate specific code snippet, cooking recipe and generating a poem about a specific topic (e.g., climate change). 
Fig.~\ref{fig:vlm_examples} shows examples demonstrating visual reasoning capabilities of our multimodal \emph{MobiLlama}-$V$. For instance, \emph{MobiLlama}-$V$ accurately describes  the atypical aspects of the image when asked to describe the given image.  

\noindent \textbf{Evaluating Large-base Model:}
As discussed earlier, we strive to develop fully transparent models for democratization of SLMs and fostering future research. To this end, we compare our \emph{large-base} 1.2B with existing fully transparent SLMs falling within the less than 2B category. 
%These existing SLMs are:  Boomer, pythia~\cite{biderman2023pythia}, Falcon-RW~\cite{refinedweb}, TinyLlama~\cite{zhang2024tinyllama}, OLMo~\cite{OLMo}, cerebras-gpt~\cite{dey2023cerebras}, Lamini-LM~\cite{lamini-lm}, opt~\cite{zhang2022opt} and gpt-neo~\cite{gpt-neo}. 
Tab.~\ref{tab:llm-comparison-1B} shows that compared to recent OLMo and TinyLlama that are pre-trained on a larger dataset of 3T tokens, our \emph{large-base} 1.2B model pre-trained on 1.2T tokens achieves favourable results with an average score of 49.06 over nine benchmarks. We hope that our \emph{large-base} model will serve as a solid baseline and help ease future research in SLM development.

\section{Conclusion}

We present a fully transparent SLM, \emph{MobiLlama},  that alleviates redundancy in the transformer block. Within \emph{MobiLlama}, we propose to utilize a shared FFN design for all the blocks within the SLM. 
%Our \emph{MobiLlama} is accurate yet efficient in terms of training cost, on-device memory and storage efficiency.
We evaluate \emph{MobiLlama}  on nine benchmarks, achieving favourable results compared to existing methods falling under less than 1B category. 
We also build a multimodal model on top of \emph{MobiLlama} SLM to demonstrate visual reasoning capabilities. 
%We hope that our \emph{MobiLlama} will help accelerate research efforts towards building fully-transparent, accurate yet efficient SLMs that bridge the gap with their resource hungry LLM counterparts. 

\noindent \textbf{Limitation and Future Direction:}  
A potential direction is to further improve \emph{MobiLlama} for enhanced context comprehension.
%and understanding subtlety of linguistic nuances. 
%Domain-specific expertise of the model can also be explored (e.g., healthcare). 
While  \emph{MobiLlama} offers a fully transparent SLM framework, a follow-up study to understand any misrepresentations and biases is desired to improve model's robustness.

\section{Acknowledgement}
The computations were enabled by the Berzelius resource provided by the Knut
and Alice Wallenberg Foundation at the National Supercomputer Centre. We thank Sahal Shaji Mullappilly and Muhammad Maaz for their support in the evaluations on mobile platform and VLM training.

\bibliography{custom}

% \pagebreak
% \newpage
\appendix

\section{Appendix}
\label{sec:appendix}

\subsection{MobiLlama-Chat}
We also developed the fine-tuned versions of MobiLlama-0.5B and MobiLlama-1.2B, namely MobiLlama-0.5B-Chat and MobiLlama-1.2B-Chat. We used the similar fine-tuning dataset and hyper-parameters as AmberChat \citep{llm360} to train our models. We use FastChat \citep{zheng2024judging} to fine-tune our models on \href{https://huggingface.co/datasets/WizardLM/WizardLM_evol_instruct_V2_196k}{WizzardLM-196K} as used by WizardLM \citep{xu2023wizardlm}. We trained our base models for 3 epochs on 8 A100s (80G), keeping the learning rate as $2 \times 10^{-5}$, gradient accumulation steps as 16, and warmup ratio as 0.04.

% \clearpage
\onecolumn
\subsection{Qualitative Examples}

\newcommand{\OutputBox}[2]{%
    \newtcolorbox{#1}{
      colback=white,
      colframe={gray!90},
      title={\textbf{#2}},
      fonttitle=\bfseries,
      boxrule=0.5mm,
      arc=3mm,
      boxsep=5pt,
      left=0pt,
      right=6pt,
      top=8pt,
      bottom=8pt,
    }
}

\OutputBox{OutputBoxA}{Reasoning:}
\begin{OutputBoxA}
\begin{itemize}[left=0pt, label={},noitemsep,topsep=0pt]
  \item \textbf{\textcolor{blue}{Human:}} What are the key benefits of practicing mindfulness meditation?
  \begin{itemize}[left=-5pt,label={},noitemsep,topsep=0pt]
  \item
  \end{itemize}
  
  \item \textbf{\textcolor{orange}{MobiLlama-0.5B:}}
  \begin{itemize}[left=0pt,label={},noitemsep,topsep=0pt]
    \item Practicing mindfulness meditation can have several benefits, including:
    \item 1. Reduced stress and anxiety: Mindfulness meditation can help reduce stress and anxiety by focusing on the present moment and letting go of distractions.
    \item 2. Improved focus and concentration: Mindfulness meditation can help improve focus and concentration by focusing on the present moment and letting go of distractions.
    \item 3. Better sleep: Mindfulness meditation can help improve sleep by reducing stress and anxiety and promoting relaxation.
    \item 4. Better mood: Mindfulness meditation can help improve mood by reducing stress and anxiety and promoting relaxation.
    \item 5. Better relationships: Mindfulness meditation can help improve relationships by reducing stress and anxiety and promoting relaxation.
  \end{itemize}
\end{itemize}
\end{OutputBoxA}

\OutputBox{OutputBoxB}{Knowledge:}
\begin{OutputBoxB}
\begin{itemize}[left=0pt, label={},noitemsep,topsep=0pt]
  \item \textbf{\textcolor{blue}{Human:}} What are some ways to reduce the carbon footprint of your home?
  \begin{itemize}[left=-5pt,label={},noitemsep,topsep=0pt]
  \item
  \end{itemize}
  
  \item \textbf{\textcolor{orange}{MobiLlama-0.5B:}}
  \begin{itemize}[left=0pt,label={},noitemsep,topsep=0pt]
    \item Reducing the carbon footprint of your home can be achieved through several ways. Here are some tips:
    \item 1. Use energy-efficient lighting and appliances.
    \item 2. Install energy-efficient windows and doors.
    \item 3. Use energy-efficient appliances and lighting.
    \item 4. Reduce the use of paper and other non-biodegradable materials.
    \item 5. Use energy-efficient appliances and lighting.
    \item 6. Reduce the use of plastic and other single-use materials.
    \item 7. Use energy-efficient appliances and lighting.
    \item 8. Reduce the use of paper and other single-use materials.
  \end{itemize}
\end{itemize}
\end{OutputBoxB}

\OutputBox{OutputBoxC}{Reasoning:}
\begin{OutputBoxC}
\begin{itemize}[left=0pt, label={},noitemsep,topsep=0pt]
  \item \textbf{\textcolor{blue}{Human:}} How do diverse ecosystems contribute to the overall health of the planet?
  \begin{itemize}[left=0pt,label={},noitemsep,topsep=0pt]
  \item
  \end{itemize}
  
  \item \textbf{\textcolor{orange}{MobiLlama-0.5B:}}
  \begin{itemize}[left=0pt,label={},noitemsep,topsep=0pt]
    \item Different cultures and traditions contribute to the diversity of the world in many ways. For example, the Chinese culture is known for its emphasis on family, tradition, and hospitality. The Japanese culture is known for its emphasis on individualism, creativity, and innovation. The Indian culture is known for its emphasis on tradition, devotion, and spirituality.
  \end{itemize}
\end{itemize}
\end{OutputBoxC}

\newcommand{\OutputBoxx}[2]{%
    \newtcolorbox{#1}{
      colback=white,
      colframe={black!100},
      title={\textbf{#2}},
      fonttitle=\bfseries,
      boxrule=0.5mm,
      arc=3mm,
      boxsep=5pt,
      left=0pt,
      right=6pt,
      top=8pt,
      bottom=8pt,
    }
}

\clearpage
\OutputBoxx{OutputBoxxA}{Story Creation:}
\begin{OutputBoxxA}
\begin{itemize}[left=0pt, label={},noitemsep,topsep=0pt]
  \item \textbf{\textcolor{blue}{Human:}} Create the opening paragraph of a suspenseful thriller: In the dimly lit alley, Detective Miller knew something was off. The air was thick with tension, and as he turned the corner, he discovered a hidden door leading to a world of 
  \begin{itemize}[left=0pt,label={},noitemsep,topsep=0pt]
  \item
  \end{itemize}
  \item \textbf{\textcolor{orange}{MobiLlama-0.8B:}}
  \begin{itemize}[left=0pt,label={},noitemsep,topsep=0pt]
    \item icy secrets that would soon be revealed in his own life."
    \item "The first thing I noticed about this book is how well-written it really felt like an actual story from beginning to end," said one reviewer on Goodreads who also gave her rating for "10 out of ten stars". Another reviewed by another author wrote:"This novel has been described many times over but never quite captured my attention until now...I'm not sure if you can put your finger on what makes me love these books so much or maybe just because they are such good reads! They have all sorts of twists along their way which keeps them fresh every time!"
  \end{itemize}
\end{itemize}
\end{OutputBoxxA}

\OutputBoxx{OutputBoxxC}{Technological Evolution:}
\begin{OutputBoxxC}
\begin{itemize}[left=0pt, label={},noitemsep,topsep=0pt]
  \item \textbf{\textcolor{blue}{Human:}} Once people considered Artificial Intelligence is something impossible to achieve, but now 
  \begin{itemize}[left=0pt,label={},noitemsep,topsep=0pt]
  \item
  \end{itemize}
  \item \textbf{\textcolor{orange}{MobiLlama-0.8B:}}
  \begin{itemize}[left=0pt,label={},noitemsep,topsep=0pt]
    \item The future of artificial intelligence: AI will be used in every aspect from medicine and healthcare through automation for manufacturing industries like robotics or even autonomous driving systems that can drive themselves without human intervention (or at least with less risk).
  \end{itemize}
\end{itemize}
\end{OutputBoxxC}

% \OutputBoxx{OutputBoxxD}{Imagination and Adventure:}
% \begin{OutputBoxxD}
% \begin{itemize}[left=0pt, label={},noitemsep,topsep=0pt]
%   \item \textbf{\textcolor{blue}{Human:}} Imagine the climax of a time-travel adventure: As the time machine hummed with energy, Sarah and Alex found themselves transported to a pivotal moment in history, where they faced a decision that could alter the course of 
%   \begin{itemize}[left=0pt,label={},noitemsep,topsep=0pt]
%   \item
%   \end{itemize}
  
%   \item \textbf{\textcolor{orange}{MobiLlama-0.8B:}}
%   \begin{itemize}[left=0pt,label={},noitemsep,topsep=0pt]
%     \item Their journey was not without its challenges; for example when their ship crash land on an alien planet inhabited by giant insects called “Buggers.” The Bugger’s are so powerful it is believed there were once two species living side by side – one from Earth (the Bugs) who lived underground while humans walked above them like plants or trees but now live below ground as bugs beneath earth surface which can be seen through windows at nighttime! They also have supernatural powers such being able to fly over people into space if you touch your hand against any object within range…and even more power than this ability comes out during battle between these creatures known as ‘Dark Beasts.’ These Dark Beast attackers come equipped with huge claws capable of ripping apart flesh leaving behind deadly venomous spikes ready to inflict painful wounds upon anyone caught up in close proximity..
%   \end{itemize}
% \end{itemize}
% \end{OutputBoxxD}

\OutputBoxx{OutputBoxxE}{Analytical Thinking:}
\begin{OutputBoxxE}
\begin{itemize}[left=0pt, label={},noitemsep,topsep=0pt]
  \item \textbf{\textcolor{blue}{Human:}} Create a concept for an album cover that represents the concept of time travel through music 
  \begin{itemize}[left=0pt,label={},noitemsep,topsep=0pt]
  \item
  \end{itemize}
  
  \item \textbf{\textcolor{orange}{MobiLlama-0.8B:}}
  \begin{itemize}[left=0pt,label={},noitemsep,topsep=0pt]
    \item and art, or something similar to this:
    \item Time Travel Through Music – A song with lyrics about how you can go back in history if only one person would listen…or maybe it’s just me! I don't know what my favorite is yet but there are so many songs out now on YouTube where people talk about going backwards into their past lives (like “Time Machine” by The Beatles) which makes sense because they were talking about getting ridiculous amounts of money from someone who was dead when we get here today; however some artists have gone further than simply saying "I want all your stuff" like Kanye West did last year...he actually went as far as creating his own musical instrument called 'The Tidal Wave'. It basically looks exactly like he does at times during his live shows except instead of being able to see himself perform inside another body while singing along, users will be transported directly onto stage via wave technology using headphones attached to them..and then afterward everyone gets together again once more!! This could also mean anything really though since most musicians do not even realize themselves performing outside of reality anymore anyway....but whatever works best for YOU!!!
  \end{itemize}
\end{itemize}
\end{OutputBoxxE}

% ________________ MobiLlama-1.2B
\newcommand{\OutputBoxxx}[2]{%
    \newtcolorbox{#1}{
      colback=white,
      colframe={black!65},
      title={\textbf{#2}},
      fonttitle=\bfseries,
      boxrule=0.5mm,
      arc=3mm,
      boxsep=5pt,
      left=0pt,
      right=6pt,
      top=8pt,
      bottom=8pt,
    }
}

\OutputBoxxx{OutputBoxxxA}{Programming and Logic Building:}
\begin{OutputBoxxxA}
\begin{itemize}[left=0pt, label={},noitemsep,topsep=0pt]
  \item \textbf{\textcolor{blue}{Human:}} Generate a Python code snippet that implements a function to calculate the Fibonacci sequence using recursion. Bonus points if you can make it more efficient using memoization.
  \begin{itemize}[left=0pt,label={},noitemsep,topsep=0pt]
  \item
  \end{itemize}
  
  \item \textbf{\textcolor{orange}{MobiLlama-1.2B:}}
  \begin{itemize}[left=0pt,label={},noitemsep,topsep=0pt]
  \item Sure, here's a Python code snippet that implements a function to calculate the Fibonacci sequence using recursion:
  \item ```
  \item \verb|def fibonacci(n):|
  \item \quad \verb|if n <= 1:|
  \item \quad \quad \verb|return n|
  \item \quad \verb|else:|
  \item \quad \quad \verb|return fibonacci(n-1) + fibonacci(n-2)| \\
    
  \item To make this function more efficient using memoization, we can store the previous values of the Fibonacci sequence in a dictionary. Here's an updated version of the function that uses memoization: \\
    
    \item \verb|def fibonacci(n, memo={}):| 
    \item \quad \verb|if n in memo:| 
    \item \quad \quad \verb|return memo[n]| 
    \item \quad \verb|elif n <= 1:| 
    \item \quad \quad \verb|return n| 
    \item \quad \verb|else:|
    \item \quad \quad \verb|memo[n] = fibonacci(n-1, memo) + fibonacci(n-2, memo)|
    \item \quad \quad \verb|return memo[n]| \\
    
    \item This function takes a number \verb|n| as input and returns the \verb|n|th Fibonacci number, or the value \verb|n| if it's already been calculated. If \verb|n| is less than or equal to 1, the function returns the value \verb|n|. Otherwise, it recursively calls itself with the previous two numbers in the sequence and adds them together. If the value of \verb|n| is already in the dictionary, the function returns the value of \verb|n| directly. Otherwise, it adds the value of \verb|n| to the dictionary and returns the value of \verb|n| with the memoized value.
    
    \item This updated function is more efficient because it avoids calculating the same value multiple times, which can be a bottleneck in recursive functions.
  \end{itemize}
\end{itemize}
\end{OutputBoxxxA}

\OutputBoxxx{OutputBoxxxB}{Innovative Thinking:}
\begin{OutputBoxxxB}
\begin{itemize}[left=0pt, label={},noitemsep,topsep=0pt]
  \item \textbf{\textcolor{blue}{Human:}} Construct a sentence that uses every letter of the alphabet at least once.
  \begin{itemize}[left=0pt,label={},noitemsep,topsep=0pt]
  \item
  \end{itemize}
  
  \item \textbf{\textcolor{orange}{MobiLlama-1.2B:}}
  \begin{itemize}[left=0pt,label={},noitemsep,topsep=0pt]
  \item I am just a 10-year-old's birthday party planner. I love organizing and planning fun and memorable events for kids. I am always up for a challenge and love to create unique and exciting experiences for my clients.
  \end{itemize}
\end{itemize}
\end{OutputBoxxxB}

\OutputBoxxx{OutputBoxxxC}{Creativity and Imagination:}
\begin{OutputBoxxxC}
\begin{itemize}[left=0pt, label={},noitemsep,topsep=0pt]
  \item \textbf{\textcolor{blue}{Human:}} Write a role-play with a tourist and a travel guide for a finctional planet. 
  \begin{itemize}[left=0pt,label={},noitemsep,topsep=0pt]
  \item
  \end{itemize}
  
  \item \textbf{\textcolor{orange}{MobiLlama-1.2B:}}
  \begin{itemize}[left=0pt,label={},noitemsep,topsep=0pt]
    \item Sure, here's a role-play:
    \item \verb|###| Tourist: Hi, I'm a tourist from Earth. I'm here to explore the finctional planet. Can you tell me more about it?
    \item \verb|###| Travel Guide: Sure, the finctional planet is a unique and fascinating place. It's a world where science and magic collide, where the laws of physics are different from our own.
    \item \verb|###| Tourist: That sounds amazing! Can you tell me more about the planet's history and culture?
    \item \verb|###| Travel Guide: Sure, the planet was created by a group of scientists who wanted to explore the limits of human knowledge. They created a simulation to test their theories, and the simulation went wrong. The simulation created a world where science and magic intertwined, and the planet became a place where people could live and work together.
    \item \verb|###| Tourist: That sounds amazing! Can you tell me more about the planet's technology and infrastructure?
    \item \verb|###| Travel Guide: Sure, the planet has advanced technology that allows people to travel through time and space. It also has a network of underground tunnels that connect different parts of the planet.
    \item \verb|###| Tourist: That sounds amazing! Can you tell me more about the planet's natural resources and resources?
    \item \verb|###| Travel Guide: Sure, the planet has a vast array of natural resources that are essential for the planet's survival. It has a unique ecosystem that is home to a variety of species, and it has a thriving economy that relies on the planet's resources.
    \item \verb|###| Tourist: That sounds amazing! Can you tell me more about the planet's history and culture?
    \item \verb|###| Travel Guide: Sure, the planet has a rich history that dates back to the beginning of time. It has a complex society that is based on a balance between science and magic.
    \item \verb|###| Tourist: That sounds amazing! Can you tell me more about the planet's future?
    \item \verb|###| Travel Guide: Sure, the planet is in the process of creating a new civilization that will be based on a fusion of science and magic. It will be a place where people can live and work together, and it will be a place where science and magic will be in harmony.
    \item \verb|###| Tourist: That sounds amazing! Can you tell me more about the planet's future?
    \item \verb|###| Travel Guide: Sure, the planet is in the process of creating a new civilization that will be based on a fusion of science and magic. It will be a place where people can live and work together, and it will be a place where science and magic will be in harmony.
  \end{itemize}
\end{itemize}
\end{OutputBoxxxC}

\end{document}